\documentclass[letterpaper]{article}
\usepackage[preprint]{preprint}
\usepackage[hyphens]{url}
\usepackage{graphicx}
\usepackage{natbib}
\usepackage{caption}
\usepackage{booktabs}
\usepackage{amsmath}
\usepackage{amssymb}
\usepackage{multirow}
\usepackage{tikz}
\usetikzlibrary{positioning}

\title{AtomCite: Verification and Correction of Supplied Page-Level Citations
in Multi-Page Documents}
\author{
  Chen Qian$^{1}$ \quad Yimeng Wang$^{1}$ \quad Yu Chen$^{2}$ \quad Lingfei Wu$^{2}$ \quad Andreas Stathopoulos$^{1}$
}
\affiliations{
  $^{1}$William \& Mary \quad $^{2}$Anytime AI \\
  \texttt{\{cqian03,ywang139,axstat\}@wm.edu} \quad \texttt{\{ychen,lwu\}@anytime-ai.com}
}

\begin{document}
\maketitle

\begin{abstract}
When answering questions based on multi-page documents, large language models are expected to cite the supporting pages. However, supplied citations are sometimes inaccurate, and current evaluations score citations at generation time or against text passages: no existing benchmark enables evaluating whether a system can verify and correct a page-level citation already attached to an answer. To address this gap, we propose \textbf{AtomCite}, an agentic framework that parses an answer into claims, checks each claim against the image of its cited page, and applies a deterministic repair policy. To evaluate it, we introduce \textbf{DocCite}, to our knowledge the first benchmark for evaluating systems that verify and correct page-level citations in document images. Built on MP-DocVQA and DUDE, it combines 928 validated injected instances with 2{,}468 candidate natural errors harvested from frontier- and efficiency-tier models, of which a two-annotator audit confirms 1{,}909 as genuine errors. Primary labels are assigned deterministically, not by LLM judges, and the human audit is a separate validation layer on top of them. Across three model families (Gemini, Claude, and GPT), AtomCite reaches around 93\% binary verification accuracy on the injected benchmark, significantly outperforming every OCR-only condition, including a compute-matched control, and exceeding every prior text-based baseline given the same OCR text. Its repair policy lifts citation precision on the injected mix from its constructed 34\% to 87--90\% while retaining over 90\% of correct claims. AtomCite also transfers: with frozen prompts and zero training, it raises the hallucination-detection scores of two open 7--8B models on five public benchmarks above the same models prompted as direct judges. Finally, a two-annotator audit shows that noise in automatic labels biases measured verifier accuracy and can reverse system rankings, so evaluations that rely only on synthetic or automatic labels risk mismeasuring verification capability.
\end{abstract}

\section{Introduction}

Large language models increasingly answer questions about long, visually complex documents \citep{mmlongbenchdoc2024benchmarking}, such as contracts, medical records, and financial filings, and are expected to justify each claim with a citation to a specific page \citep{magesh2025hallucination}. Supplied citations matter: a reviewer who trusts a wrong page number may act on unverified content, and a fabricated claim with a plausible-looking citation is more dangerous than an uncited one. Courts have sanctioned attorneys for filing AI-fabricated authority, with over a thousand documented decisions worldwide \citep{mata2023avianca,charlotin2026hallucination}. Yet the reliability of supplied page-level citations is largely unmeasured. Attribution work evaluates citations to \emph{retrieved text passages} in open-domain settings \citep{rashkin2023measuring,liu2023evaluating}; multi-page document QA benchmarks record which page holds the evidence only as ground truth for \emph{answering} \citep{mpdocvqa2023hierarchical,dude2023document}; and recent benchmarks that pair citations with document images \citep{citevqa2026benchmarking,mmlongcite2025benchmark} enable scoring those citations only as the answer is \emph{generated}. Post-hoc repair exists for text settings: CiteFix repoints passage-level RAG citations and RARR revises unsupported claims \citep{citefix2025enhancing,rarr2023researching}. To our knowledge, no existing benchmark, in the attribution, document-QA, or citation-repair line above, supports evaluating whether a system can verify and correct page-level citations already attached to an answer over a multi-page document.

We study the closed-world version of this problem: given an answer, its claims, the cited pages, and the source document (as page images with OCR text), verify each (claim, cited page) pair and repair the citation layer. The closed world reflects the deployment setting of document-grounded assistants, isolates citation verification from retrieval quality, and bounds all claims in this paper.

Verification here is multimodal. On text, prompted frontier judges are competitive detectors \citep{minicheck2024efficient,hu2025decomposition}. On document images, we find prompting insufficient at every tested scale. The same claim may be supported by a table cell, a signature block, or a stamp OCR renders poorly or not at all; conversely, a page whose OCR text \emph{seems} to support a claim may contradict it visually. In our experiments, neither layout-aware OCR nor a compute-matched text condition closes this gap.

\begin{figure*}[t]
\centering
\includegraphics[width=0.94\textwidth]{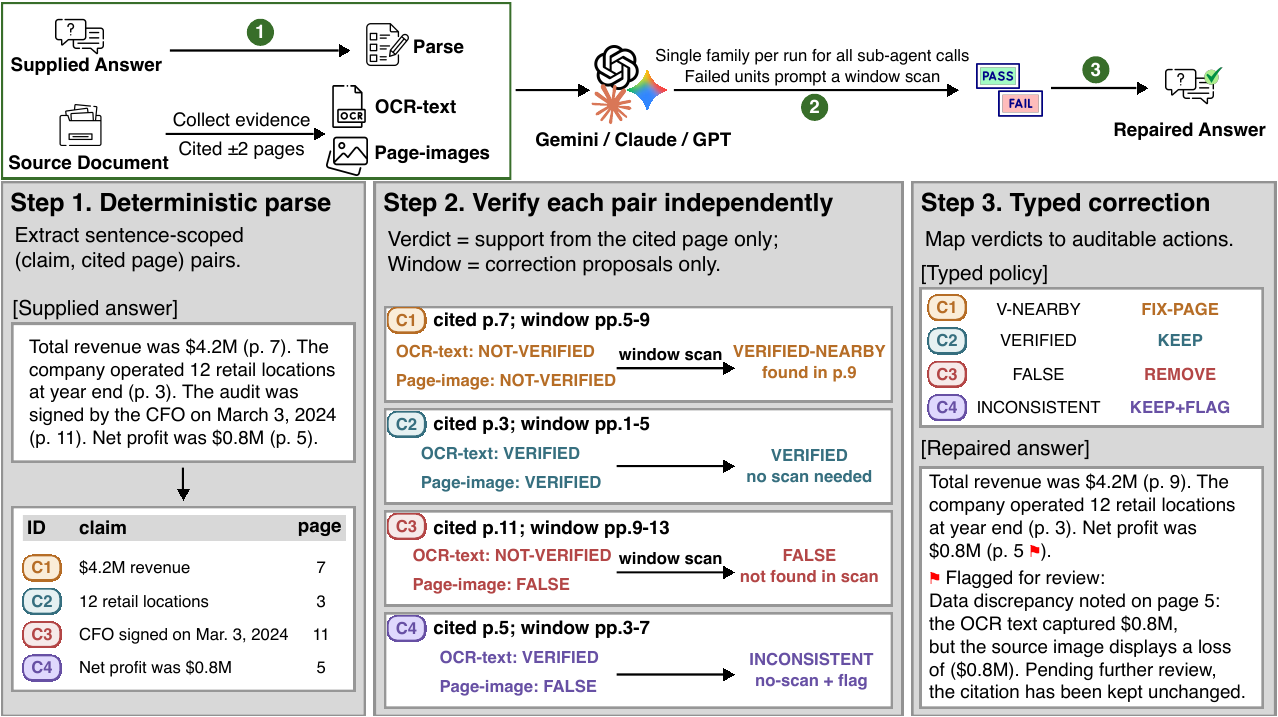}
\caption{The AtomCite pipeline on a four-claim answer, the five stages
grouped into parse, verify, and correct. Each parsed (claim, cited
page) pair is checked by text and image sub-agents on the cited page
only; window pages contribute correction proposals, never
verdicts. The example covers all four actions: a true claim cited to
the wrong page is repointed (\textsc{fix-page}), a verified claim
kept, a claim supported nowhere removed, and a hard text--image
conflict (\textsc{inconsistent}) kept and flagged, never
deleted.}
\label{fig:pipeline}
\end{figure*}

We make three primary contributions. \textbf{(1) AtomCite}, an agentic verify-and-correct framework for supplied page-level citations (Fig.~\ref{fig:pipeline}). It parses the answer into atomic units (claim + cited page pairs) and checks each claim with parallel grounding sub-agents that read the OCR text and the image of the cited page. The verdict for a claim uses the \emph{cited page only}; surrounding pages contribute correction proposals, never verdicts. A typed policy maps verdicts to actions (\textsc{keep}, \textsc{fix-page}, \textsc{remove}). \textbf{(2) DocCite}, to our knowledge the first benchmark for evaluating systems that verify and correct supplied page-level citations in multi-page document images. It has two components: \textbf{DocCite-Syn}, 928 controlled injected instances whose labels are individually validated by deterministic document-wide scans, and \textbf{DocCite-Nat}, 2{,}468 naturally occurring candidate citation errors harvested from frontier- and efficiency-tier model generations over MP-DocVQA and DUDE documents, with audit-validated labels. Ground truth is two-layered: deterministic page-localization (no LLM judge assigns primary test labels) plus a separately reported human-audited semantic layer. The audited layer is necessary: we show noise in automatic natural-error labels biases measured verifier accuracy (flat OCR appears to beat image grounding under them). \textbf{(3)} Evidence that the results are not an artifact of one setup: the modality and architecture findings replicate across three model families and two OCR engines, and the framework transfers with frozen prompts and zero training to five public hallucination benchmarks, improving two open 7--8B backbones.

Across three model families, image-aware verification beats every OCR-only condition. AtomCite reaches $93.7\%$ binary accuracy with Gemini, $93.0\%$ with Claude, and $93.0\%$ with GPT; every OCR rung of the conditions ladder falls significantly below it in every family. A compute-matched OCR control does not close the gap in any family, and a sequential single-transcript chain at matched compute also trails significantly in all three. The correction policy raises citation precision from $34.3\%$ to $87$--$90\%$ while retaining $\approx$$92\%$ of correct claims across all three families, versus $50.3\%$ and $36.8\%$ for reimplemented CiteFix and adapted RARR. Under the deterministic Layer-1 labels, natural errors appear systematically harder to detect than injected ones ($70$--$90\%$ across conditions vs.\ $92$--$97\%$); a two-annotator audit shows much of that gap is label noise rather than verifier failure.\looseness-1
\section{Related Work}
\label{sec:related}

\paragraph{Answer attribution and citation evaluation.} A large body of work asks whether generated answers are supported by cited evidence. The AIS framework \citep{rashkin2023measuring} formalized human judgments of attribution, later automated as AutoAIS \citep{bohnet2022attributed}; ALCE \citep{alce2023enabling} established benchmark protocols for the citation quality of generated answers, and audits of generative search engines found supplied citations frequently fail to support their statements \citep{liu2023evaluating}. Subsequent work refines the unit of analysis via decomposition into atomic claims \citep{factscore2023fine,veriscore2024evaluating}, builds automatic evaluators of citation quality \citep{citeeval2025principle,attributionbench2024how}, and contrasts generation-time with post-hoc citation \citep{saxena2025generation}. In high-stakes legal settings, fabricated or unsupported citations are documented in deployed retrieval-augmented tools \citep{dahl2024large,magesh2025hallucination}, and fabricated scholarly references motivate reference-verification benchmarks \citep{citeaudit2026cited}. Across this line the evidence is free text (passages, long contexts, or bibliographic records), and the object of study is generating or scoring citations. None of these efforts centers on verifying, and then repairing, page-level citations already supplied with an answer over a document rendered as images.\looseness-1

\paragraph{Multi-page document QA.} Document VQA \citep{docvqa2021dataset} has been extended to multi-page settings \citep{mpdocvqa2023hierarchical,dude2023document}, with long-document suites that annotate evidence pages \citep{mmlongbenchdoc2024benchmarking}. These resources supply questions with gold evidence-page annotations, but the page label serves as ground truth for \emph{answering}; the benchmarks pose neither verification nor correction of a citation attached to a claim. We build on this line rather than compete with it: our testbeds derive documents, questions, and evidence-page annotations from MP-DocVQA and DUDE, which jointly provide native page imagery and redistribution-friendly licenses. Concurrent work has begun to couple citations with document images and long multimodal contexts \citep{citevqa2026benchmarking,docscope2026benchmarking,mmlongcite2025benchmark,multattnattrib2026training}. These efforts evaluate or produce citations at \emph{generation} time, leaving supplied-citation verification and correction unaddressed.\looseness-1

\paragraph{Verification and correction systems.} Grounding detectors score claim--evidence consistency with NLI-style or purpose-built models \citep{summac2022revisiting,alignscore2023evaluating,minicheck2024efficient, hhem2024open,lettucedetect2025hallucination} or sampling-based consistency \citep{selfcheckgpt2023zero}. ContextCite \citep{contextcite2024attributing} and SelfCheckGPT presume access to the generator (its internals or its samples), which a post-hoc auditor lacks; among NLI-style detectors we benchmark the four recent ones that report superseding SummaC, plus a general-purpose DeBERTa NLI model \citep{laurer2024less}. On the correction side, RARR \citep{rarr2023researching} retrieves evidence and revises unsupported statements, and CiteFix \citep{citefix2025enhancing} repairs RAG citations with lexical and semantic matching. These detectors and correctors consume clean text spans, so we adapt them to our setting as baselines under explicitly disclosed input-faithfulness tiers (Sec.~\ref{sec:experiments}). Text-grounded detection also has established public benchmarks with human labels, including RAGTruth \citep{niu2024ragtruth}, TofuEval \citep{tang2024tofueval}, FaithBench \citep{bao2025faithbench}, HalluMix \citep{emery2025hallumix}, and VeriGray \citep{ding2025verigray}. Finally, because LLM evaluators favor their own generations \citep{panickssery2024llm}, our primary test labels are deterministic and are not assigned by LLM judges at evaluation time. Concurrent benchmarks evaluate adjacent problems (multimodal claim--evidence consistency \citep{musciclaims2025multimodal,m2verify2026large}; legal citation retrieval and reliability \citep{sglegalcite2026principle,legalcitebench2026evaluating}), but none evaluates systems that verify and \emph{correct} supplied page-level citations in multi-page document images; to our knowledge, ours is the first benchmark centered on that task.\looseness-1
\section{Testbeds and Ground Truth}
\label{sec:testbeds}

\paragraph{Task.} A \emph{citation instance} is a pair (atomic claim, cited page). Given the source document as page images with per-page OCR, a system decides whether the cited page supports the claim (\textsc{supported} vs.\ not): the deployment question is whether the citation stands as supplied, and this binary cited-page support is the primary metric. For correction, non-supported verdicts are further typed by whether the document supports the claim elsewhere (\textsc{wrong-location}, repairable) or nowhere (\textsc{unsupported}, removable). Supported-claim recall is reported alongside it. Systems may abstain; abstentions are excluded from accuracy and reported per condition.\looseness-1

\paragraph{Two-layer ground truth.} Primary test labels are never assigned by an LLM judge at evaluation time. \emph{Layer 1} is deterministic page-localization, a proxy for support: a citation is correctly localized iff the cited page belongs to the instance's gold page set. Evidence can legitimately appear on several pages. Gold sets are therefore expanded from the source benchmarks' single evidence pages by a document-wide fuzzy-matching scan, adding 505 evidence pages across 2{,}058 QA pairs. Scan parameters were frozen on the development split, and the expansion is human-audited. \emph{Layer 2} is semantic claim support: a stratified human calibration of Layer-1 labels on $\geq$300 instances (including $\geq$150 abstractive claims). The Layer-1$\leftrightarrow$Layer-2 discrepancy rate and its confidence interval are reported alongside every headline table.\looseness-1

\paragraph{Documents.} We build on MP-DocVQA and DUDE documents spanning forms, letters, reports, tables, and mixed print--handwriting scans. A frozen seeded selection fixes three disjoint pools: an injection corpus of 240 documents (180 MP-DocVQA, 60 DUDE; 3{,}398 pages), a 150-document natural-generation pool (100/50), and a 30-document development split for parameter freezing. MP-DocVQA pages are OCR'd with a uniform Tesseract~5.5.2 pass; DUDE pages use the dataset-shipped Azure OCR. We pick Tesseract \citep{tesseractocr} as the primary engine because its pinned version keeps the text layer fully reproducible; a stronger learned engine (MinerU~2.5; \citealp{wang2026mineru25propushinglimitsdatacentric}) replicates the whole OCR ladder in Sec.~\ref{sec:experiments}. A layout-aware serialization (region tags, reading order, table cell separators) is derived for the OCR-only conditions of Sec.~\ref{sec:experiments}.\looseness-1

\paragraph{DocCite-Nat: natural errors.} Six generator configurations (three model families $\times$ frontier/efficiency tiers) answer document questions with required page citations. A deterministic marker parser extracts (claim, cited page) instances from the raw answers (sentence-scoped claims; page lists, ranges, footnotes), so the pool exercises answer parsing end-to-end. The pool is built as a funnel, each step reported: Layer-1 labeling of all generations yields \textbf{2{,}468} answer-bearing candidate errors (a claim carrying the gold answer, cited to a page outside the expanded gold set). A two-annotator blind audit then adjudicates every candidate on which any verifier arm disagreed with the Layer-1 label (two census waves: 608 disputed by a family's AtomCite verdict, 306 by only the OCR-only or staged arms). The remaining $1{,}554$ candidates, disputed by no arm, keep the Layer-1 error label, and a seeded 50-item spot-check tests that region. Partial support counts as an error: a citation must fully support its claim. The audit thus partitions the pool ($608{+}306{+}1{,}554$) into \textbf{1{,}909} audit-validated errors ($240$ of them partial-support), \textbf{450} confirmed-supported claims (Layer-1 label noise, reported as a finding), and \textbf{109} annotator-mixed items excluded from all denominators. Detection is scored against the validated errors; false alarms against the confirmed-supported claims (Sec.~\ref{sec:experiments}). Candidate-error prevalence ranges from $44.7\%$ to $61.2\%$ across generators, and cross-family probes bound self-preference effects.\looseness-1

\paragraph{DocCite-Syn: validated injection.} For controlled measurement we inject 1{,}000 instances over the same documents at the natural-error mix: clean (318), wrong-page (402), partially supported (127), fabricated (92), and value-perturbed (61). Every non-clean template is validated by a deterministic document-wide scan with quote-anchored value extraction and normalization. The scan guarantees, e.g., that a ``perturbed'' value appears nowhere in the document and that a ``wrong page'' does not carry the claim. Instances failing their scan are discarded: of 1{,}000 constructed, 32 were found accidentally supported and 40 carried duplicate evidence, leaving the 928 we score (318 clean, 362 wrong-page, 113 partially supported, 82 fabricated, 53 value-perturbed). Labels follow a single template-first rule set shared in code by the generator, the scorer, and the audit trace.\looseness-1
\section{The AtomCite Framework}
\label{sec:framework}

AtomCite processes an answer in five stages (Fig.~\ref{fig:pipeline}); the design constraint throughout is that \emph{uncertainty must never silently delete content or grant it false authority}.

The name reflects the unit of verification: each parsed claim is checked whole as one atomic unit against its cited page (a finer decomposition variant is ablated later).\looseness-1

\paragraph{(1) Parse.} Deterministic extraction of claims and citation markers (inline page references, ranges, multi-cite groups, footnotes), producing (claim, cited page) instances.

\paragraph{(2) Evidence.} For each citation: the cited page (image + OCR), and a context window of $\pm 2$ pages capped at five evidence pages. The cited page is the only admissible verdict evidence; window pages feed correction proposals exclusively. This separation is enforced in the type system, not by prompting.\looseness-1

\paragraph{(3) Ground.} Per claim, parallel sub-agents check the cited page through two channels: a text channel over the page OCR and an image channel over the page image. Units that fail on the cited page trigger a window scan (per channel) whose only output is a candidate supporting page.

\paragraph{(4) Synthesize.} Channel results merge under typed rules: a hard conflict (one channel \textsc{verified}, the other \textsc{false}) yields \textsc{inconsistent}, which is excluded from scoring and reported per condition (Statistics); on softer disagreements the image channel is authoritative on document cases, since OCR omissions are the dominant error source. Channel verdicts aggregate conservatively (any \textsc{false} dominates), with a \textsc{verified-nearby} verdict when the claim verifies on a window page instead of the cited page. For scoring, \textsc{verified} maps to \textsc{supported} (labels of Sec.~\ref{sec:testbeds}), \textsc{verified-nearby} to \textsc{wrong-location}, and the remaining verdicts to \textsc{unsupported}. A flag-only \emph{consistency check} then compares verdicts across the answer's citations and marks mutually contradictory ones \textsc{inconsistent} (excluded downstream); it can flag but never alter a per-citation verdict.\looseness-1

\paragraph{(5) Correct.} A typed policy maps verdicts to actions: \textsc{verified}$\to$\textsc{keep}; \textsc{verified-nearby}$\to$\textsc{fix-page} (repoint to the supported page); \textsc{not-verified}/\textsc{false}$\to$\textsc{remove} (whole-claim removal at sentence boundaries); \textsc{inconsistent}/abstain$\to$\textsc{keep} with a flag. The headline configuration is this conservative, fully label-scorable mode; a full mode adds \textsc{rewrite-claim} and is evaluated in the human-calibrated stratum. We report ($\Delta$precision, supported-claim retention) so deletion-heavy strategies cannot pass as improvements.\looseness-1

All stages run identically across model families; per-condition compute (calls, tokens, list-price cost) is logged per instance and reported with every comparison.
\section{Experiments}
\label{sec:experiments}

\paragraph{Conditions as ablations.} The conditions ladder is the paper's ablation study: each condition changes exactly one ingredient, so each pairwise comparison isolates one design question. Codes group by axis: A-series = OCR-text input, B-series = prompting, C = the full framework. The \emph{modality axis} asks what the image channel contributes. \textbf{(A1)} runs the identical framework with the page image replaced by flat OCR text (remove the image, keep everything else). \textbf{(A2)} upgrades to layout-aware OCR serialization (does lost \emph{layout} explain the gap?). \textbf{(A3)} additionally spends the image condition's budget on four deterministic text passes (entities, numbers, page-location, tables) aggregated by pre-specified conjunction (does budget explain it?).\looseness-1

The \emph{prompting axis} asks whether a strong or staged prompt matches the framework. \textbf{(B1)} replaces the entire pipeline with one direct verification call (is any structure needed?). \textbf{(B2)} adds a structured claim-by-claim checklist to that single call. \textbf{(B3)} executes the framework's stages as one growing single-transcript chain over the same evidence at matched call count (same compute and staged reasoning, without the agentic decomposition into typed parallel sub-tasks). \textbf{(C)} is the full framework: claim-level verification with both channels.\looseness-1

Both compute-matched controls are calibrated to a fixed reference budget ($3.4\times$ layout-OCR list cost; $3.5$ calls/item) that upper-bounds (C)'s own cost, so neither control lacks budget. Further pipeline ablations appear with the results. All LLM conditions run the family's \emph{frontier} backbone under pinned identifiers (\texttt{gemini-3.1-\allowbreak pro-preview}, \texttt{claude-\allowbreak opus-4-8}, \texttt{gpt-5.6-sol}), with greedy decoding where the API exposes temperature (0.0) and provider defaults otherwise, and a fixed 4{,}096-token output cap; the efficiency identifiers (\texttt{gemini-3-\allowbreak flash-preview}, \texttt{claude-\allowbreak sonnet-5}, \texttt{gpt-5.6-luna}) enter as DocCite-Nat generators and in the tier-robustness check of Appendix~\ref{app:extra}. OCR text follows the per-source engines of Sec.~\ref{sec:testbeds} (Tesseract~5.5.2 / Azure); the whole A~ladder is replicated under a second, stronger engine (MinerU~2.5; Table~\ref{tab:conditions}). Baseline tiers are disclosed as T1 (as intended), T2 (adapted), T3 (reimplemented).

\paragraph{Statistics.} Paired cluster bootstrap on condition differences (document clusters, $B{=}10{,}000$); significance in Table~\ref{tab:diffs} is Holm-corrected across its 18 comparisons (family-wise $\alpha{=}0.05$), and other comparisons report per-comparison 95\% CIs. Abstention is zero in every main-matrix cell after retry exhaustion. Excluded verdicts (\textsc{inconsistent} or unparseable output) are at most 13 of 928 per cell ($\leq$$1.4\%$), nearly all (C)'s own consistency flags. Scoring every exclusion as an error, the worst case for (C), keeps 16 of Table~\ref{tab:diffs}'s 18 differences Holm-significant. The two exceptions, (C)$-$(A3$'$) Claude and (C)$-$(B3) GPT, keep positive point estimates. Run-to-run variance is small where measured (three (C) repeats span $0.33$pp) and other cells are single runs. Text-quoted differences are paired on commonly scored instances and computed before rounding, so they can deviate slightly from rounded table entries.

\begin{table}[t]
\centering
\small
\setlength{\tabcolsep}{2.5pt}
\begin{tabular}{@{}lcccccc@{}}
\toprule
& \multicolumn{2}{c}{Gemini} & \multicolumn{2}{c}{Claude} & \multicolumn{2}{c}{GPT} \\
\cmidrule(lr){2-3}\cmidrule(lr){4-5}\cmidrule(lr){6-7}
Condition & Acc & Sup.\,R & Acc & Sup.\,R & Acc & Sup.\,R \\
\midrule
(A1) Tesseract flat      & 84.5 & 59.7 & 84.8 & 63.5 & 83.7 & 58.5 \\
(A2) Tesseract layout   & 85.2 & 62.3 & 84.2 & 63.8 & 85.2 & 61.9 \\
(A3) Tesseract matched & 86.2 & 62.6 & 86.7 & 71.7 & 87.8 & 69.2 \\
(A1$'$) MinerU flat     & 87.7 & 69.8 & 88.4 & 70.8 & 89.0 & 73.6 \\
(A2$'$) MinerU layout   & 88.5 & 72.0 & 89.1 & 74.5 & 88.1 & 71.4 \\
(A3$'$) MinerU matched  & 88.7 & 72.6 & 90.5 & 78.6 & 89.0 & 75.8 \\
(B1) direct              & 87.9 & 81.4 & 81.8 & \textbf{90.6} & 88.4 & \textbf{94.7} \\
(B2) structured         & 88.1 & 80.8 & 88.6 & 88.4 & 89.3 & 88.4 \\
(B3) seq.\ matched      & 87.5 & 75.2 & 89.1 & 83.3 & 90.6 & 85.8 \\
(C) \textbf{AtomCite}   & \textbf{93.7} & \textbf{90.4} & \textbf{93.0} & 90.4 & \textbf{93.0} & 91.0 \\
\bottomrule
\end{tabular}
\caption{DocCite-Syn conditions ladder ($n{=}928$; \%). Acc = binary
cited-page support accuracy (primary metric); Sup.\,R = recall on
truth-supported instances. (C) is claim-level AtomCite (its decomposition
variant is ablated with the results).
Bold = per-family best among image-grounded conditions;
the two (B1) Sup.\,R cells above (C) reflect lenient acceptance, not
sharper discrimination (see Architecture vs.\ Prompting);
paired differences with CIs in Table~\ref{tab:diffs}. The matched
control (A3) reaches its score with near-zero wrong-location recall.}
\label{tab:conditions}
\end{table}

\begin{table}[t]
\centering
\small
\setlength{\tabcolsep}{2pt}
\begin{tabular}{@{}lccc@{}}
\toprule
Difference & Gemini & Claude & GPT \\
\midrule
\multicolumn{4}{@{}l}{\emph{Modality axis: what does the image channel contribute?}} \\
(C)$-$(A1$'$)        & +5.9 [3.8,8.0] & +4.5 [2.4,6.5] & +3.9 [2.0,5.8] \\
(C)$-$(A2$'$)       & +5.1 [3.2,7.1] & +3.7 [1.8,5.6] & +4.8 [2.8,6.8] \\
(C)$-$(A3$'$)      & +4.9 [3.0,6.9]  & +2.3 [0.3,4.2]  & +3.9 [2.1,5.8] \\
\midrule
\multicolumn{4}{@{}l}{\emph{Prompting axis: does staged prompting close the gap?}} \\
(C)$-$(B1)        & +5.9 [3.4,8.5] & +11.2 [8.7,13.8] & +4.8 [2.6,6.9] \\
(C)$-$(B2)       & +5.5 [3.4,7.6] & +4.4 [2.4,6.6]  & +3.8 [1.9,5.7] \\
(C)$-$(B3)       & +6.2 [3.7,8.7] & +3.8 [1.9,5.7]  & +2.5 [0.5,4.4] \\
\bottomrule
\end{tabular}
\caption{Key paired differences in binary accuracy (pp; doc-cluster
bootstrap, $B{=}10^4$, 95\% CIs). Condition codes as in
Table~\ref{tab:conditions}; the modality rows use the stronger MinerU
engine (corpus-engine gaps are larger). All 18 differences are
Holm-significant (CIs per-comparison).}
\label{tab:diffs}
\end{table}

\subsection{Image-Aware vs.\ OCR-Only Verification}
Removing the image channel (Table~\ref{tab:conditions}) costs $2$--$6$pp under MinerU (Table~\ref{tab:diffs}), $5$--$9$pp under the corpus engines, and the loss concentrates in supported-claim recall (Gemini: $90.4\%$ with the image vs.\ at best $72.6\%$ without): on scan-dominated corpora, OCR-only verification rejects correctly cited claims.\looseness-1

Two alternative explanations are tested by controls; neither closes the gap. \emph{Lost layout information?} Layout-aware serialization (A2) recovers at most $1.5$pp over flat OCR ($+0.8$ Gemini, $-0.7$ Claude, $+1.5$ GPT; only GPT's gain is significant). \emph{Less compute?} The matched control (A3), which spends the image budget on four extra deterministic text passes, moves a non-significant $+1.0$pp beyond (A2) (Gemini; see below for the replication families). A tripling of text-side tokens across (A1)$\to$(A2)$\to$(A3) buys $+1.7$pp on Gemini, $+1.9$pp on Claude, and $+4.1$pp on GPT in total (Cross-Family Replication below), against same-engine image gaps of $+5.1$ to $+8.9$pp: budget alone is an unlikely explanation.\looseness-1

\subsection{Architecture vs.\ Prompting}
The prompting axis of Table~\ref{tab:diffs} reads top to bottom as progressively stronger prompting, and every staged arm trails the framework significantly in every family. Even the sequential matched-compute chain (the framework's own stages in one growing transcript, at the larger matched budget) remains $2.5$--$6.2$pp below (C). Direct prompting's two high cells (GPT's $94.7$ and Claude's $90.6$ Sup.\,R) reflect lenient acceptance, not sharper discrimination: (B1) verifies $14.9\%$ (GPT) and $22.8\%$ (Claude) of unsupported instances, two-and-a-half to four times (C)'s $5$--$6\%$. Staged-prompting recovery is family-dependent: Claude and GPT climb monotonically across (B1)$\to$(B3), while Gemini's three staged arms are flat within half a point, leaving the residual gap largest for Gemini and smallest for GPT ($2.5$pp).\looseness-1

\subsection{Cross-Family Replication}
Both replication families reproduce the modality result (modality block of Table~\ref{tab:diffs}): every OCR rung loses to (C) in every family, every confidence interval excludes zero, and the intervals overlap across families: the same effect at comparable magnitude (overlapping CIs), with the same supported-recall signature. The decomposition variant replicates the pattern at slightly smaller margins: it trails (C) by $1.2$--$1.8$pp yet still clears flat OCR by $6.3$--$7.5$pp in every family (per-comparison CIs exclude zero; Ablations below). Extra text-side compute buys a little in some families (at most $+2.6$pp, Claude and GPT), and in none does it approach the image gap.\looseness-1

\paragraph{OCR-engine robustness.} Because the A rungs inherit the corpus OCR engines, we re-OCR every page of the injection corpus with MinerU~2.5 and re-run the whole A~ladder in every family with the serialization formats held fixed (rows (A1$'$)--(A3$'$) of Table~\ref{tab:conditions}). The stronger engine lifts the OCR-only rungs by $+1.2$ to $+5.3$pp binary and $+6.6$ to $+15.1$pp supported recall, yet every gap to (C) stays significant; the modality rows of Table~\ref{tab:diffs} report these MinerU gaps as the headline numbers, the conservative choice (corpus-engine gaps are larger). The modality gap narrows but never closes.\looseness-1

\subsection{Ablations}
Beyond the conditions ladder, four pipeline ablations change one internal design choice each, holding everything else fixed. \emph{Decomposition}: adding a decomposition stage that splits each claim into finer-grained facts and aggregates by strict conjunction \emph{lowers} accuracy by $1.2$--$1.8$pp (significant for Gemini and Claude; the $1.2$pp gap for GPT does not reach significance, $p{=}0.075$). Per-fact grounding noise compounds on multi-fact claims. AtomCite therefore verifies claims whole, offering decomposition only for fine-grained auditing. \emph{Correction window} (under this variant; Gemini): a sweep from $\pm1$ to $\pm5$ pages leaves binary accuracy exactly unchanged ($91.9\%$ at every width): verdicts are decided by the cited page alone, wider windows degrade repair precision, and the pre-registered $\pm2$ default stands. \emph{Consistency stage} (same setup): disabling the multi-citation coherence check gives $91.8\%$ ($-0.1$pp, ns): it guards coherence, not per-citation verdicts. Both probe the typed evidence separation, identical with and without decomposition. \emph{Channel arbitration}: replacing the image-authoritative rule for soft channel disagreements with a symmetric pessimistic meet collapses (C) from $93.7\%$ to $86.3\%$ ($-7.4$pp $[5.1,9.8]$) and supported recall from $90.4\%$ to $65.0\%$, near OCR levels. The rule was fixed at design time, before any condition was scored. Ground truth backs it: the image channel is right in $77$--$87\%$ of decisive soft disagreements.\looseness-1

\begin{table}[t]
\centering
\small
\setlength{\tabcolsep}{2pt}
\begin{tabular}{@{}llcc@{}}
\toprule
System & Tier & Acc & Sup.\,R \\
\midrule
CiteEval-Auto (Gem/Cl/GPT) & T2 & 89.0/88.2/88.9 & --- \\
HHEM            & T1 & 83.9 & 64.5 \\
MiniCheck       & T1 & 82.9 & 57.5 \\
LettuceDetect   & T1 & 81.7 & 52.5 \\
AlignScore      & T1 & 76.0 & 33.0 \\
DeBERTa-NLI     & T1 & 72.8 & 81.4 \\
ALCE-AutoAIS    & T2 & 70.3 & --- \\
\midrule
\textbf{AtomCite (C)}, same backbones & --- & \textbf{93.7/93.0/93.0} & 90.4--91.0 \\
\bottomrule
\end{tabular}
\caption{Text-Baseline Tier on DocCite-Syn ($n{=}928$; \%; detector
operating point $0.5$). T1 = specialized detectors run as intended;
T2 = adapted protocols. Same-family paired differences,
(C)$-$CiteEval-Auto: $+4.5$ $[2.0,7.1]$ Gemini, $+4.9$ $[2.2,7.7]$
Claude, $+4.1$ $[1.7,6.7]$ GPT, all significant.}
\label{tab:baselines}
\end{table}

\subsection{Text-Baseline Tier}
Table~\ref{tab:baselines} compares AtomCite against five specialized support detectors and two protocol baselines. Every text baseline trails the framework; the strongest, CiteEval-Auto, shares the framework's own backbone and still trails by $4.1$--$4.9$pp. Because backbone strength is a confound, we re-run CiteEval-Auto on all three frontier backbones and test the \emph{same-family} paired difference. It favors (C) in every family, all significant (caption of Table~\ref{tab:baselines}), so the advantage does not reduce to backbone strength. T1 detectors are compared against the OCR conditions, their fair reference class. Empty-OCR pages (image-only) score zero for all text systems.\looseness-1

\begin{table}[t]
\centering
\small
\setlength{\tabcolsep}{2pt}
\begin{tabular}{@{}lccccccccc@{}}
\toprule
& \multicolumn{3}{c}{Gemini} & \multicolumn{3}{c}{Claude} & \multicolumn{3}{c}{GPT} \\
\cmidrule(lr){2-4}\cmidrule(lr){5-7}\cmidrule(lr){8-10}
Condition & Det & FA & Acc & Det & FA & Acc & Det & FA & Acc \\
\midrule
\textbf{AtomCite (C)}     & 97.7 & \textbf{5.5} & \textbf{97.1} & 97.3 & \textbf{5.4} & \textbf{96.8} & 98.6 & \textbf{12.3} & \textbf{96.5} \\
Flat OCR (A1)    & \textbf{99.7} & 43.8 & 91.4 & \textbf{99.8} & 48.2 & 90.7 & \textbf{99.7} & 49.3 & 90.3 \\
Seq.\ chain (B3) & 91.1 & 16.9 & 89.6 & 87.2 & 10.0 & 87.7 & 96.3 & 12.4 & 94.6 \\
\bottomrule
\end{tabular}
\caption{DocCite-Nat against audit-validated labels (\%; best per
column in bold). Partial support counts as an error. Det = detection
of the $1{,}909$ audit-validated errors; FA = false-alarm rate on the $450$
human-confirmed supported claims; Acc = overall accuracy on the
$2{,}359$ audited instances. Rates are over the audited candidate
pool, not all supplied citations. Balanced accuracy: C
$96.1$/$96.0$/$93.1$, A1 $78.0$/$75.8$/$75.2$, B3
$87.1$/$88.6$/$91.9$ (Gemini/Claude/GPT).}
\label{tab:naturals}
\end{table}

\subsection{Natural Errors}
A fixed-protocol human audit evaluates the naturals: two annotators with document-analysis experience, independent and \emph{blind}. Sheets show only the claim and the cited-page image (no verdicts or labels; order shuffled per annotator), and disagreements remain unadjudicated. They labeled four sets: a census of every instance where any family's claim-level (C) verdict disagreed with the Layer-1 label ($n{=}608$, agreement $0.89$), a census of the region where only the OCR-only or staged arms disagreed ($n{=}306$, agreement $0.837$), a stratified 320-item calibration sample ($\kappa{=}0.87$), and a seeded 50-item spot-check of the undisputed region ($0\%$ noise, Wilson 95\% CI $[0,7.3]$). Consensus rules: both-supported $=$ label noise, both-partial or both-unsupported $=$ validated error, annotator-mixed $=$ excluded. With the $1{,}554$ undisputed candidates kept at Layer-1, the partition covers $2{,}468$.

Verifier conditions are (C), flat OCR (A1), and the sequential chain (B3); the intermediate rungs are isolated on DocCite-Syn. Table~\ref{tab:naturals} reports performance against the resulting human labels. AtomCite (C) detects $97.3$--$98.6\%$ of audit-validated errors at the lowest false-alarm rate, and every miss is a partial-support item: on fully unsupported claims its detection is $100\%$. Flat OCR misses almost nothing ($99.7$--$99.8\%$) only by false-alarming on nearly half the genuinely supported claims, and the matched sequential chain (B3) misses $4$--$13\%$ of errors, including fully unsupported ones. The two census waves test whether the audit favors (C): label noise is $72.9\%$ among the instances (C) disputed but $2.3\%$ where only (A1) or (B3) did. Two findings generalize beyond our system. First, \emph{automatic natural-error labels understate verifiers}: under the raw Layer-1 labels flat OCR appeared \emph{better} than image-grounded verification, an inversion driven by false alarms agreeing with the noise. Second, the pre-registered label-downgrade rule fired for the natural strata: calibration-sample discrepancy is $65.3\%$ $[55.7,73.9]$ on extractive-natural and $36.7\%$ $[29.4,44.6]$ on abstractive items, so those labels are reported only with this audit. The injected strata passed ($7.2\%$ $[3.1,15.9]$): the labels behind the primary DocCite-Syn results pass human calibration.\looseness-1

\begin{table}[t]
\centering
\small
\setlength{\tabcolsep}{2pt}
\begin{tabular}{@{}lcccc@{}}
\toprule
System & Precision & Retention & Coverage & Fix-reach \\
\midrule
\textbf{AtomCite (Gemini)} & 34.3 $\to$ \textbf{89.9} & 91.8 & 95.4 & 68.9 \\
\textbf{AtomCite (Claude)} & 34.3 $\to$ \textbf{87.7} & 91.8 & 94.1 & 72.9 \\
\textbf{AtomCite (GPT)}    & 34.3 $\to$ \textbf{87.0} & 92.5 & 94.1 & 67.7 \\
CiteFix (T3, reimpl.) & 34.3 $\to$ 50.3 & 100.0$^{\dagger}$ & 54.9 & \textbf{80.5} \\
RARR (T2, adapted) & 34.3 $\to$ 36.8 & 94.0 & 15.7 & 0.0 \\
\bottomrule
\end{tabular}
\caption{Correction on DocCite-Syn (conservative mode; \%). Precision =
citation precision before $\to$ after correction; Retention = correct
claims kept; Coverage = error instances acted on; Fix-reach = exact gold-page
recovery on window-reachable wrong-location citations.
$^{\dagger}$CiteFix is replacement-only (cannot remove): its
$100\%$ retention is uninformative.}
\label{tab:correction}
\end{table}

\subsection{Correction}
AtomCite lifts citation precision from $34.3\%$, the synthetic mix's constructed prevalence, to $87$--$90\%$ while retaining $91.8$--$92.5\%$ of correct claims across all backbones (Table~\ref{tab:correction}). The baselines fail in complementary ways. RARR's edit-only agreement gate rarely fires on wrong-page citations and its edits damage $6\%$ of correct claims. CiteFix repoints well (Table~\ref{tab:correction}) but cannot remove fabricated claims. Repointing and removal fail independently: a correction system needs both. On the audited naturals the policy acts on $97.2$--$98.6\%$ of validated errors, retains $88.0$--$94.7\%$ of confirmed-supported claims, and lands $50.8$--$58.7\%$ of \textsc{fix-page} targets. Fix-reach conditions on reachability: $69.3\%$ of wrong-page instances hold their gold page within $\pm2$; the rest fall to \textsc{remove} (Ethics).\looseness-1

\begin{table}[t]
\centering
\small
\setlength{\tabcolsep}{2.6pt}
\begin{tabular}{@{}lcccc@{}}
\toprule
& \multicolumn{2}{c}{Qwen2.5-7B} & \multicolumn{2}{c}{Llama-3-8B} \\
\cmidrule(lr){2-3}\cmidrule(lr){4-5}
Benchmark (metric) & direct & \textbf{AtomCite} & direct & \textbf{AtomCite} \\
\midrule
RAGTruth (F1)        & 49.5 & \textbf{71.2} & 56.4 & \textbf{66.2} \\
FaithBench (bAcc)    & 50.8 & \textbf{57.0} & 52.8 & \textbf{56.9} \\
HalluMix (acc)       & 76.3 & \textbf{80.0} & 58.0 & \textbf{59.7} \\
TofuEval (bAcc)      & 67.7 & \textbf{69.1} & 54.3 & \textbf{64.2} \\
VeriGray resp.\ (F1) & 29.8 & \textbf{44.6} & 39.7 & \textbf{54.7} \\
\bottomrule
\end{tabular}
\caption{Transfer to five public hallucination benchmarks (\%), under
official test sets and metrics
($n{=}2{,}700$/$750$/$6{,}500$/$534$/$329$): F1 = response-level F1,
bAcc = balanced accuracy, acc = accuracy. The direct arm is a
same-backbone single-call judge; AtomCite is the framework
in text mode.}
\label{tab:transfer}
\end{table}

\subsection{Transfer to Public Hallucination Benchmarks}
\label{sec:transfer}
To test external validity we transplant AtomCite to five public hallucination benchmarks, keeping its staged verification design and dropping the image channel and correction policy. The direct arm is a single-call judge, the (B1) analogue. Prompts are frozen once for all five benchmarks. Backbones are Qwen2.5-7B \citep{qwen2025qwen25} and Llama-3-8B \citep{grattafiori2024llama3}, matching the 7--8B scale of open detectors evaluated by HalluMix and VeriGray. VeriGray is scored response-level from its sentence labels: a summary is unfaithful when it contains a fabricated or contradicting sentence.\looseness-1

AtomCite improves on the direct judge in all ten cells of Table~\ref{tab:transfer}. Reference numbers are quoted from the cited papers' tables on the same test sets and metrics, so no baseline is weakened. On RAGTruth Qwen reaches 71.2 F1 ($+21.7$ $[18.4, 25.0]$), above every reported zero-training system (best 63.4) and below detectors fine-tuned on its training split (79.2--83.9). On FaithBench it reaches 57.0, within 0.7 of the best published number (57.7, GPT-4-Turbo). On HalluMix it lands within a point of fine-tuned 7--8B specialists (80.0 against 80.8). On TofuEval both backbones clear the open 13B--33B evaluators reported there ($50$--$60$). Every specialist above needs task-specific fine-tuning. AtomCite needs none.\looseness-1
\section{Discussion, Limitations, and Ethics}
\label{sec:discussion}

\paragraph{What the results support.} In the closed-world supplied-citation setting, image-aware verification outperforms every text-only condition in all three families: neither extra text passes nor sequential inference recovers the gap. A conservative typed policy converts verification into large precision gains without deletion gaming. We claim no cross-domain text-vs-image generality, no open-retrieval setting, and no model-insensitivity beyond consistent within-family modality gains. On cost, (C) spends $2$--$3\times$ the layout-OCR list price, so the ladder gives deployments a cost--accuracy choice: (A2) at a third the cost where a missed false citation is tolerable, the full verifier where it is not.\looseness-1

\paragraph{Limitations.} (i) Layer-1 ground truth is page-localization rather than entailment. The human audit corrects this for the naturals: both disagreement regions are censused. (ii) Gold-set expansion has finite recall: missed duplicate evidence deflates stronger systems (conservative for our claims). (iii) Baseline tiers T2/T3 are adaptations, not the original systems (deviations in Appendix~\ref{app:extra}). (iv) Our documents are English business/legal images: other scripts, born-digital PDFs, and non-page granularities are out of scope.\looseness-1

\paragraph{Ethics and impact.} A citation verifier is itself an authority: a false \textsc{remove} deletes truthful content, and \textsc{fix-page} can repoint a fabricated claim to a plausible page. Uncertainty never deletes, actions are typed and auditable, retention is reported alongside precision, and high-stakes deployments should keep a human in the loop for \textsc{remove}. The benchmark redistributes only what licenses permit (DUDE CC-BY; identifiers and scripts for MP-DocVQA), and code, prompts, labels, audit adjudications, and run ledgers are released with the paper.\looseness-1

\section{Conclusion}
We introduced AtomCite, which verifies claims against cited page images and repairs failures with a typed policy, and DocCite, a benchmark for the task. Across three families and two OCR engines, the page image beat every OCR-only and staged condition, including both compute-matched controls. Correction raised precision from $34.3\%$ to $87$--$90\%$ at $\approx$$92\%$ retention. AtomCite improved two open 7--8B backbones on five hallucination benchmarks. A two-annotator audit showed that noisy automatic labels bias measured
verifier accuracy enough to reverse system rankings, so DocCite ships
human-audited labels.\looseness-2

{\small
\bibliography{references}}

\clearpage
\appendix
\setcounter{table}{0}
\setcounter{figure}{0}
\renewcommand{\thetable}{A\arabic{table}}
\renewcommand{\thefigure}{A\arabic{figure}}
\section*{Appendix: Supplementary Material}

\section{Testbed Construction Detail}
\label{app:construction}

\paragraph{Document selection.} Seeded, frozen selections (seed 20260716) fix three disjoint document pools, chosen before any system output was scored: the injection corpus of 240 documents (180 MP-DocVQA, expanded from a 100-document pilot pool, plus 60 DUDE; 3{,}398 pages), a 150-document natural-generation pool (100 MP-DocVQA, 50 DUDE; 2{,}068 pages), and a 30-document development split (425 pages) on which the gold-expansion scan parameters were frozen. No document appears in more than one pool. Every selected page was rendered and OCR-processed with a single engine version; per-page TSV output additionally feeds the layout-aware serialization (region blocks with coordinates, reading order, \texttt{|}-separated table cells) used by conditions (A2)/(A3).

\paragraph{Gold-set expansion.} Source benchmarks give one evidence page per QA pair; duplicate evidence (running headers, repeated fields, summary pages) makes single-page gold sets wrong for localization scoring. We scan every document for near-matches of each QA's evidence string: sliding window of 10 tokens, fuzzy threshold $0.85$, minimum match length 12, and a distinctiveness guard discarding matches whose text occurs on more than 3 pages (boilerplate). Parameters were frozen on a development split before test scoring; the expansion added 505 evidence pages across 2{,}058 QA pairs. A random $n{=}50$ human audit of expansion decisions is reported with the release. The human expansion audit (all 76 added pages from the $n{=}50$ QA sample, two annotators, $89.5\%$ agreement) finds $60.5\%$ $[49.3,70.8]$ of added pages carry fully equivalent evidence ($71.1\%$ under an either-annotator reading): expansion trades some gold-set precision for recall, and both error directions are covered by the Layer-2 calibration above.

\paragraph{DocCite-Syn templates.} Each instance derives from a QA pair with a verified answer-bearing claim. \emph{clean}: claim + correct page. \emph{wrong\_page}: claim + shifted page; shifts that land inside the (expanded) gold set are re-shifted at generation. \emph{perturb}: an entity/number in the claim is replaced by a type-compatible alternative (same interrogative class and surface shape); a document-wide scan verifies the perturbed value appears nowhere. \emph{partial}: a supported base claim with an appended unsupported statement whose value the scan verifies absent. \emph{fabricated}: a claim foreign to the document, scan-verified. Scan value extraction is quote-anchored with digit/comma normalization; instances failing their scan are discarded (of 1{,}000 constructed, 32 accidentally supported and 40 duplicate-evidence instances are removed; the 928 validated instances are scored). The discards by template are wrong\_page 40, partial 14, fabricated 10, and perturb 8, leaving a final mix of 318 clean, 362 wrong-page, 113 partial, 82 fabricated, and 53 perturbed. Among the 362 wrong-page instances, 251 ($69.3\%$) have their nearest gold page within the $\pm2$ correction window (237 at distance 1, 14 at distance 2); the 111 beyond-window instances are unreachable by \textsc{fix-page} and fall to \textsc{remove} under the policy. Label derivation is a single template-first rule set shared in code by generator, scorer, and audit traces.

\paragraph{Natural-error harvest.} Six generator configurations (Claude, Gemini, GPT families $\times$ frontier/efficiency tiers) produced cited answers for the selected QA pairs. Layer-1 labeling against expanded gold sets yielded 2{,}468 answer-bearing candidate errors across all six configurations, far exceeding the pre-registered $N{\geq}180$ powering threshold (an interim two-family harvest of 433 already cleared it during construction). The two-annotator audit (two census waves plus spot-check) then partitioned the candidates into 1{,}909 audit-validated errors (240 of them partial-support, counted as errors under the strict metric), 450 confirmed-supported claims, and 109 annotator-mixed items (see the funnel figure below). Matched correctly-localized generations serve as negatives in the stratified calibration sample.

\paragraph{Candidate-error prevalence per generator.} Table~\ref{tab:prevalence} reports, for each generator configuration, the number of answer-bearing cited claims and the fraction Layer-1 labels as candidate errors. Prevalence spans $44.7$--$61.2\%$ and neither tier is uniformly cleaner: the Claude frontier generator errs more often than its efficiency sibling, while the Gemini and GPT frontier generators err less. GPT generators produce far fewer answer-bearing cited claims than the other families. Counts are computed on the current generation snapshot, which reproduces $2{,}392$ of the $2{,}468$ pool rows; the remaining $76$ originate from superseded generation passes retained for pool-index stability and lack denominators, so they are excluded here.

\begin{table}[h]
\centering
\small
\begin{tabular}{lrrr}
\toprule
Generator & AB claims & Cand.\ errors & Prev.\ (\%) \\
\midrule
Gemini frontier    &   954 & 426 & 44.7 \\
Gemini efficiency  & 1{,}076 & 556 & 51.7 \\
Claude frontier    & 1{,}037 & 590 & 56.9 \\
Claude efficiency  & 1{,}098 & 601 & 54.7 \\
GPT frontier       &   284 & 134 & 47.2 \\
GPT efficiency     &   139 &  85 & 61.2 \\
\bottomrule
\end{tabular}
\caption{Layer-1 candidate-error prevalence by generator
(AB = answer-bearing cited claims). Candidate status is the Layer-1
localization label; the two-annotator human audit validates
$77.4\%$ of pooled candidates as genuine errors (partial support
counted as error).}
\label{tab:prevalence}
\end{table}

\paragraph{Annotation ethics.} The audited documents are public-benchmark pages; annotation involved no crowd work and no personal data beyond what the source benchmarks release.

\paragraph{Audit trail.} Before launch, the pipeline passed a 20-case micro-audit and an independent two-author hand-trace of scorer arithmetic (which caught and fixed two real defects: a duplicated label-derivation path and a wrong-page shift landing inside a multi-page gold set --- both now regression-tested); all amendments are date-logged in the released execution record.
\section{Additional Results}
\label{app:extra}

\paragraph{Condition codes.} A1/A2/A3 = flat/layout/compute-matched OCR input; B1/B2/B3 = direct/structured/sequential-chain prompting; C = the full framework. C-atom denotes C's atomic-decomposition variant. Released artifacts key these conditions as \texttt{d}/\texttt{d2}/\texttt{d2s} (A-series), \texttt{a}/\texttt{a2}/\texttt{a3} (B-series), \texttt{c}, and \texttt{e} (C-atom).

\paragraph{Text-pool testbed (AtomCite-Legal).} On a 500-run legal QA benchmark with bounded text-context pools, strict citation precision of unassisted frontier generation is $31.4\%$ ($37.8\%$ lenient) --- establishing that supplied citations are unreliable even in the easy, text-only case. Verification conditions on the 400-question test split ($n{=}2{,}687$ claim-citation pairs per cell; $\leq 5$ abstentions per GPT cell, zero elsewhere): Gemini (C-atom) $92.3\%$ binary vs.\ (C) $92.7\%$; Claude (C-atom) $90.9\%$ vs.\ (C) $92.3\%$; GPT (C-atom) $91.3\%$ vs.\ (C) $92.4\%$. In all three families, on short single-source legal claims the claim-level default matches or slightly exceeds the decomposition variant on binary accuracy. The small spread across conditions here motivates the document-image setting, where the architectural gaps appear.

\paragraph{Detector threshold sweeps.} Body results fix the detector operating point at $0.5$. Sweeps (appendix diagnostics only, never operating points): MiniCheck peaks at $84.9\%$ ($\tau{=}0.1$), HHEM at $83.9\%$ ($\tau{=}0.5$), LettuceDetect $83.2\%$ ($\tau{=}0.1$), AlignScore $80.0\%$ ($\tau{=}0.3$), DeBERTa-NLI $77.7\%$ ($\tau{=}0.9$). Even at test-set-optimal thresholds every detector remains below the default (C) by $\geq 8$pp.

\paragraph{Baseline adaptations (T2/T3).} T1 detectors run as released (operating point $0.5$; sweeps above). CiteFix (T3) is reimplemented from its paper, which released no code: each claim is one factual point carrying one citation, the candidate pool is the evidence window, and we run both the keyword and the LLM matching variants, reporting the LLM matcher (both recorded in the release); the fine-tuned-BERTScore variant requires the authors' private training data and is omitted. CiteFix by design only \emph{replaces} citations, so it cannot remove fabricated claims. RARR (T2) ports the official chain and prompts verbatim, swaps the deprecated \texttt{text-davinci-003} backbone for our pinned runners, and skips web retrieval: the cited page's OCR text serves as the evidence for the agreement gate and editor, with the upstream edit-distance guard at its default. RARR is edit-only --- it never proposes a page repoint --- so its exact-page recovery of $0.0$ (Table~\ref{tab:correction}) is structural, not a tuning artifact, and its agreement gate rarely fires on wrong-page citations whose text the page still supports. AutoAIS (T2) reduces in our single-citation setting to its degenerate case of one NLI call (cited page as premise, claim as hypothesis); the 11B TRUE-NLI checkpoint is substituted by the DeBERTa NLI already run in the detector tier (a pre-registered swap). CiteEval-Auto (T2) runs the native context-attribution, citation-editing, and rating chain with verbatim prompts through the same pinned backbones as the framework (the same-family pairing of Table~\ref{tab:baselines}). Full deviation catalogues accompany the code release.

\paragraph{Per-condition compute.} Per-item averages from the run ledger (list prices, uncached calls): (C-atom) 3.5 calls / 7{,}488 tokens / \$0.030; (B3) 3.5 / 8{,}283 / \$0.029; (B1) 1.0 / 5{,}664 / \$0.018; (B2) 1.0 / 5{,}870 / \$0.020; (A2) 1.3 / 1{,}582 / \$0.009; (A3) 4.0 / 4{,}681 / \$0.028. The decomposition variant costs $3.4\times$ the layout-OCR condition per item at list prices. The default (C), measured on its two cache-disabled repeats (its original run shared cached grounding calls with (C-atom) on single-atom claims), averages 2.8 calls / 5{,}951 tokens / \$0.024 per item --- cheaper than the variant, as the structure implies: one call fewer per item (no decomposition) and one grounding pass per channel per claim rather than per atom. Whether the image premium justifies its cost is deployment-dependent, and the conditions ladder is exactly the menu for that decision.

\begin{table}[t]
\centering
\small
\setlength{\tabcolsep}{4pt}
\begin{tabular}{@{}lccc@{}}
\toprule
Condition & Binary acc.\ (\%) & \$/item & Calls/item \\
\midrule
(C) \textbf{AtomCite}    & \textbf{93.7} & 0.024$^{*}$ & 2.8 \\
(C-atom) & 91.9 & 0.030 & 3.5 \\
(B3) seq.\ chain & 87.5 & 0.029 & 3.5 \\
(A3) Tesseract matched & 86.2 & 0.028 & 4.0 \\
(B2) checklist  & 88.1 & 0.020 & 1.0 \\
(B1) direct      & 87.9 & 0.018 & 1.0 \\
(A2) Tesseract layout & 85.2 & 0.009 & 1.3 \\
\bottomrule
\end{tabular}
\caption{Cost-effectiveness on DocCite-Syn (Gemini backbone; list
prices, uncached calls, per-item ledger averages). $^{*}$(C) measured on
its two cache-disabled repeats (run-to-run variance paragraph); its
original run shared cached grounding calls with (C-atom) on single-atom
claims. Call/token structure is identical across families;
absolute prices differ by provider list price.}
\label{tab:cost}
\end{table}

\paragraph{Natural-error audit trail.} Under the deterministic Layer-1 labels (before the human audit), detection of the 2{,}468 candidate natural errors is: (A1) $89.1$/$90.1$/$90.0\%$, (C-atom) $81.6$/$84.1$/$83.7\%$, (C) $80.0$/$79.4$/$82.4\%$, (B3) $76.2$/$70.3$/$78.9\%$ (Gemini/Claude/GPT) --- every image-grounded condition below flat OCR, the opposite ordering from DocCite-Syn, and the anomaly that triggered the audit. The stratified calibration sample ($n{=}320$; $150$ abstractive; $60$-item oversampled leniency stratum) gave $\kappa{=}0.87$, false-support prevalence $30.3\%$ $[24.2,37.0]$ among Layer-1 natural-error labels, false-wrong-location $3.0\%$, and $52/60$ leniency-stratum items resolved in the framework's favor --- corroborated by an automated probe finding $66\%$ of the framework's rescue quotes verbatim in the flat OCR the (A1) condition read and missed. The first census wave then adjudicated all 608 disputed instances: $72.9\%$ both-supported (label noise), $17.6\%$ partial support, $9.5\%$ annotator-mixed (excluded from all denominators), and zero consensus-unsupported --- no family missed a single clean error. Under the strict metric (partial support counts as an error), (C)'s detection is $97.3$--$98.6\%$ and every one of its misses is a partial-support item. False alarms are measured on the census-confirmed supported set, and the undisputed region is covered by the spot-check and the stratified sample. Cross-family probes (verifier $\times$ generator, condition (C)): Gemini-generated errors are hardest for every verifier ($73.8$--$76.9\%$ under Layer-1 labels); residual self-leniency is small and inconsistent (Claude $-4$ to $-6$pp on its own errors, GPT none), so difficulty is generator-specific, not verifier-specific.

\paragraph{Second census wave (A1/B3-only disagreement region).} The first wave covers every instance where a (C) verdict disagreed with the Layer-1 label. Its complement, instances where \emph{only} the flat-OCR or sequential-chain arms disagreed, contains $306$ instances after removing the $50$ spot-check items ($299$ B3-only, $6$ both, $1$ A1-only; region rule and seeds frozen in the released artifacts). The two annotators ran the identical blind protocol on all of them (raw agreement $0.837$). Consensus buckets: $7$ label noise, $139$ fully unsupported, $110$ partial support (validated errors under the strict metric), $50$ annotator-mixed. Noise among the $306$ disputed labels is $2.3\%$ (Wilson 95\% $[1.1,4.6]$; $2.7\%$ $[1.3,5.5]$ of consensus-decided items, $4.8\%$ $[2.3,9.6]$ under a lenient reading that also excludes partials), versus $72.9\%$ $[69.2,76.2]$ of the $608$ (C)-disputed labels under the same definition: label noise concentrates where (C) disputed the labels, so the census design does not manufacture (C)'s advantage. The headline partition ($1{,}909/450/109$) folds in both waves. Orderings are metric-robust: under the lenient reading (C) keeps the lowest false-alarm rate ($5.5$/$5.4$/$12.3$) and (B3)'s detection rises to $92.1$/$93.7$/$98.3$ because many of its apparent misses are partial-support items, while it still misses errors (C) does not.

\paragraph{Spot-check noise propagation.} The undisputed region ($1{,}553$ retained errors) is covered by the seeded 50-item spot-check (0\% noise, Wilson 95\% CI $[0,7.3]$). Propagating the upper bound reassigns at most 113 of those errors to supported. Because every arm flags them, the shift is symmetric: detection moves by at most $0.8$pp for any system, worst-case false-alarm rates rise for all systems together (C $24.4$--$29.9$, B3 $28.1$--$33.6$, A1 $55.1$--$59.5$), and no cross-condition ordering reverses. (C) keeps the lowest false-alarm rate and the highest balanced accuracy ($84.3$--$86.5$ vs.\ B3 $78.5$--$83.0$, A1 $70.1$--$72.3$) in every family under this worst case.

\paragraph{Per-condition exclusions and the excluded-as-error recompute.} Abstention after retry exhaustion is zero in every main-matrix cell. Excluded verdicts (\textsc{inconsistent} or unparseable output) range from 0 to 13 of 928 per cell; the largest are GPT (C-atom) 13, Claude (C-atom) 12, and the (C) cells at 6--8; all Tesseract OCR-only cells are 0 (one MinerU cell has 1, one prompting cell 2). Because these flags are almost entirely (C)'s own, we recompute all 18 headline paired differences scoring every excluded verdict as an error, the worst case for (C). Sixteen of 18 remain Holm-significant with nearly unchanged estimates. The two weakest comparisons keep positive point estimates but lose significance: (C)$-$(A3$'$) Claude moves from $+2.3$ $[0.3,4.2]$ to $+1.9$ $[0.0,4.0]$ (Holm $p{=}0.11$) and (C)$-$(B3) GPT from $+2.5$ $[0.5,4.4]$ to $+1.8$ $[-0.2,3.9]$ (Holm $p{=}0.11$). No ordering reverses under this treatment (\texttt{itt\_channel.json}).

\paragraph{Channel-disagreement audit (basis of the image-authoritative rule).} On the (C) runs over DocCite-Syn, the two grounding channels agree on $704$--$741$ of 928 instances per family; hard conflicts (\textsc{verified} vs.\ \textsc{false}) number 6--8 and become \textsc{inconsistent}. Among the decisive soft disagreements (exactly one channel returns \textsc{verified}): Gemini 91, Claude 120, GPT 100. Of these, $86$--$99\%$ are text-negative/image-positive (90/91, 103/120, 97/100), the OCR-omission signature, and scan-derived ground truth sides with the image channel in $86.8$/$76.7$/$80.0\%$ (Gemini/Claude/GPT). Following the text channel instead would be correct in $13$--$23\%$. This audit is post-hoc explanation, not rule fitting: the rule was fixed at design time from the qualitative observation that OCR omits stamps, handwriting, and table structure, was frozen before any full condition was scored (decision log in the released ledger), and an end-to-end ablation measures its effect ($-7.4$pp when replaced by a symmetric meet).

\paragraph{Correction on natural errors.} Applying the conservative correction policy to DocCite-Nat and scoring against the human-audited partition: on the 450 human-confirmed supported claims, \textsc{keep} retention is $94.7\%$ (Gemini), $94.7\%$ (Claude), and $88.0\%$ (GPT); on the 1{,}909 validated errors, the policy acts (\textsc{fix-page} or \textsc{remove}) on $97.2$--$98.6\%$ (the remainder are flagged \textsc{keep}s and partial-support items the verifier judged verified). Of the \textsc{fix-page} proposals, $50.8$/$58.7$/$51.6\%$ (Gemini/Claude/GPT) land inside the expanded gold set. The last figure is a conservative lower bound: the expansion audit puts the gold sets' own equivalent-evidence precision at $60.5\%$ with limited recall, so a proposal outside the set is not necessarily wrong; item-level human scoring of fix targets is left to future work.

\paragraph{Run-to-run variance.} Running the flagship (C) condition three times on the Gemini backbone --- the original run plus two repeats with caching disabled --- gives binary accuracy $93.70$/$93.36$/$93.37\%$ --- range $0.33$pp, standard deviation $0.17$pp --- with only 5 of 916 common items ($0.5\%$) changing their binary outcome in any repeat (the common set drops items excluded --- \textsc{inconsistent} or unparseable --- in any of the three repeats). Run-to-run noise is an order of magnitude smaller than every effect we report (the smallest significant paired difference has a CI lower bound of $+0.3$pp).

\paragraph{Source stratification.} Splitting DocCite-Syn by source corpus, condition (C): MP-DocVQA $94.7$--$95.4\%$ binary across the families ($n{=}716$--$718$); DUDE $86.8$--$87.8\%$ ($n{=}204$--$205$) --- DUDE's born-digital/scanned mix and longer documents are uniformly harder, and the family ranking is unchanged within each source.

\paragraph{OCR-engine robustness (full breakdown).} The corpus OCR engines are Tesseract~5.5.2 (MP-DocVQA) and the dataset-shipped Azure OCR (DUDE). To test whether the modality gap depends on engine quality, we re-OCR every page of the injection corpus with MinerU~2.5 (\texttt{MinerU2.5-Pro-2604-1.2B}, vLLM serving, greedy decoding) and rebuild both A-serializations from its output with the formats held fixed: (A1) reading-order flat text with tables flattened row-wise; (A2) the same \texttt{[region $x,y$ $w{\times}h$]} block headers as the corpus-engine serializer, coordinates normalized to 0--1000. (A3) then reuses its standard construction on top of the MinerU passes. Table~\ref{tab:mineru-full} gives the full breakdown. MinerU lifts supported recall by $+6.6$ to $+15.1$pp on every rung in every family --- it genuinely recovers evidence the corpus engines miss --- yet all nine paired gaps to (C) remain significant under a Holm correction (these MinerU gaps are the modality rows of Table~\ref{tab:diffs}). The corpus-engine gaps, also all Holm-significant, are larger: (C)$-$(A1)/(A2)/(A3) $= {+}8.6/{+}7.8/{+}6.8$ Gemini, ${+}7.8/{+}8.6/{+}6.0$ Claude, ${+}8.9/{+}7.5/{+}5.1$ GPT (95\% CI lower bounds $2.3$--$6.3$).

\begin{table}[t]
\centering
\small
\setlength{\tabcolsep}{4pt}
\begin{tabular}{@{}llcccc@{}}
\toprule
& Condition & Bin & Sup.\,R & Gap to (C) & 95\% CI \\
\midrule
\multirow{3}{*}{\rotatebox{90}{Gem.}}
& (A1) flat    & 87.7 & 69.8 & $+5.9$ & $[3.8,8.0]$ \\
& (A2) layout  & 88.5 & 72.0 & $+5.1$ & $[3.2,7.1]$ \\
& (A3) matched & 88.7 & 72.6 & $+4.9$ & $[3.0,6.9]$ \\
\midrule
\multirow{3}{*}{\rotatebox{90}{Cl.}}
& (A1) flat    & 88.4 & 70.8 & $+4.5$ & $[2.4,6.5]$ \\
& (A2) layout  & 89.1 & 74.5 & $+3.7$ & $[1.8,5.6]$ \\
& (A3) matched & 90.5 & 78.6 & $+2.3$ & $[0.3,4.2]$ \\
\midrule
\multirow{3}{*}{\rotatebox{90}{GPT}}
& (A1) flat    & 89.0 & 73.6 & $+3.9$ & $[2.0,5.8]$ \\
& (A2) layout  & 88.1 & 71.4 & $+4.8$ & $[2.8,6.8]$ \\
& (A3) matched & 89.0 & 75.8 & $+3.9$ & $[2.1,5.8]$ \\
\bottomrule
\end{tabular}
\caption{Full MinerU~2.5 replication of the A~ladder ($n{=}928$; \%).
Gap = paired difference (C)$-$rung, doc-cluster bootstrap,
$B{=}10^4$; all nine are Holm-significant ($\alpha{=}0.05$, family of
nine). Against the corpus-engine ladder, the stronger engine lifts
every rung (supported recall by $+8$--$15$pp) without closing the
image gap in any family.}
\label{tab:mineru-full}
\end{table}

\paragraph{Full metric breakdown.} Table~\ref{tab:conditions-full} reports wrong-location recall for every condition and family, alongside binary accuracy and supported recall. The key pattern: the matched OCR control (A3) trades localization away entirely (WL~R ${\leq}1.4$) for its binary score, while both AtomCite configurations keep wrong-location recall in the $50$--$58$ range --- the capability the correction policy consumes.
\begin{table}[t]
\centering
\small
\setlength{\tabcolsep}{5pt}
\begin{tabular}{@{}llccc@{}}
\toprule
& Condition & Binary & Sup.\ R & WL R \\
\midrule
\multirow{8}{*}{\rotatebox{90}{Gemini}}
& (A1) Tesseract flat      & 84.5 & 59.7 & 56.9 \\
& (A2) Tesseract layout   & 85.2 & 62.3 & 57.7 \\
& (A3) Tesseract matched & 86.2 & 62.6 & \phantom{0}0.6 \\
& (B1) direct              & 87.9 & 81.4 & 59.9 \\
& (B2) structured         & 88.1 & 80.8 & 55.8 \\
& (B3) seq.\ matched      & 87.5 & 75.2 & 58.6 \\
& (C-atom) \textbf{AtomCite}$_{\text{+dec}}$ & 91.9 & 84.6 & 50.0 \\
& (C) \textbf{AtomCite}   & 93.7 & 90.4 & 52.8 \\
\midrule
\multirow{8}{*}{\rotatebox{90}{Claude}}
& (A1) Tesseract flat      & 84.8 & 63.5 & 62.7 \\
& (A2) Tesseract layout   & 84.2 & 63.8 & 61.9 \\
& (A3) Tesseract matched & 86.7 & 71.7 & \phantom{0}1.4 \\
& (B1) direct              & 81.8 & 90.6 & 55.5 \\
& (B2) structured         & 88.6 & 88.4 & 54.7 \\
& (B3) seq.\ matched      & 89.1 & 83.3 & 61.3 \\
& (C-atom) \textbf{AtomCite}$_{\text{+dec}}$ & 92.0 & 86.5 & 55.6 \\
& (C) \textbf{AtomCite}   & 93.0 & 90.4 & 57.2 \\
\midrule
\multirow{8}{*}{\rotatebox{90}{GPT}}
& (A1) Tesseract flat      & 83.7 & 58.5 & 60.5 \\
& (A2) Tesseract layout   & 85.2 & 61.9 & 58.3 \\
& (A3) Tesseract matched & 87.8 & 69.2 & \phantom{0}0.3 \\
& (B1) direct              & 88.4 & 94.7 & 58.6 \\
& (B2) structured         & 89.3 & 88.4 & 47.0 \\
& (B3) seq.\ matched      & 90.6 & 85.8 & 61.3 \\
& (C-atom) \textbf{AtomCite}$_{\text{+dec}}$ & 91.9 & 85.1 & 51.2 \\
& (C) \textbf{AtomCite}   & 93.0 & 91.0 & 53.3 \\
\bottomrule
\end{tabular}
\caption{Full metric breakdown of the conditions ladder ($n{=}928$; \%).
WL~R = recall on
truth-wrong-location instances. The compute-matched OCR control (A3)
reaches its binary score with no localization ability: its
wrong-location recall collapses to
${\leq}1.4$ in every family --- its binary gains come entirely from
coarse supported/unsupported calls. Wrong-location recognition is the
capability the correction policy builds on.}
\label{tab:conditions-full}
\end{table}

\paragraph{Correction-window sweep.} Widening or narrowing the correction window leaves verification untouched --- binary accuracy is exactly $91.9\%$ at every width from $\pm1$ to $\pm5$, five-point confirmation that verdicts are decided by the cited page alone --- but correction quality degrades monotonically with width: repair precision falls from $91.8\%$ ($\pm1$) to $86.9\%$ ($\pm5$) and exact gold-page recovery collapses beyond $\pm3$ (Table~\ref{tab:window}). Wider windows admit more candidate pages, and repair proposals scatter onto plausible-but-wrong neighbors; the distribution of wrong-location offsets (65.5\% within $\pm1$, 69.3\% within $\pm2$, 92.5\% within $\pm5$) means wider windows raise the \emph{reachable} ceiling while lowering realized repair quality. Note the evidence-page cap (5 pages) truncates the nominal $\pm4$/$\pm5$ windows on longer documents with a low-page bias, so those two rows underestimate what an uncapped wide window could reach --- but the monotone precision decline already visible under the cap, together with rising token cost, is what the window decision trades against. The pre-registered $\pm2$ default sits at the balance point.

\begin{table}[t]
\centering
\small
\begin{tabular}{@{}lcccccc@{}}
\toprule
Window & Bin & Prec. & Ret. & Reach & All & Pages \\
\midrule
$\pm1$ & 91.9 & \textbf{91.8} & 85.8 & \textbf{69.2} & 45.3 & 1.77 \\
$\pm2$ (default) & 91.9 & 89.9 & 86.5 & 65.3 & 45.3 & 3.33 \\
$\pm3$ & 91.9 & 88.6 & 86.5 & 62.9 & \textbf{45.9} & 3.77 \\
$\pm4$ & 91.9 & 88.0 & 86.2 & 47.1 & 36.5 & 4.00 \\
$\pm5$ & 91.9 & 86.9 & 87.7 & 38.5 & 35.6 & 4.00 \\
\bottomrule
\end{tabular}
\caption{Correction-window sweep (Gemini, $n{=}928$; \%; run under
the decomposition-variant configuration --- verification is
window-invariant in both configurations, and the correction trade-off
direction carries over). Bin =
verification binary accuracy (invariant by design); Prec.\ =
post-correction citation precision; Reach = exact recovery on
wrong-location citations reachable within each window; All = exact
recovery over all wrong-location citations; Pages = mean
effective window size after document-boundary and evidence-cap
truncation. Marginal re-run token cost rises monotonically with width
(from $\sim$2.2k/item at $\pm1$ to $\sim$4.2k at $\pm5$).}
\label{tab:window}
\end{table}

\paragraph{Correction detail.} Conservative-mode action distributions and per-template outcomes; FIX\_PAGE candidates come only from the $\pm2$ window, so $111$ of $362$ wrong-location instances are structurally unreachable (gold page outside every window) --- reachable-subset recovery is the honest denominator reported in the body alongside the overall rate.

\paragraph{The decomposition variant (C-atom).} A single optional stage splits each claim into finer-grained checkable facts (\emph{atoms}) using one pinned decomposer model; each atom is grounded separately and the claim verdict is the strict conjunction of atom verdicts. Wherever decomposition appears in a comparison the same pinned, cached decomposition is shared by every system, eliminating atom-alignment bias. On DocCite-Syn the variant reaches $91.9$/$92.0$/$91.9\%$ binary (Gemini/Claude/GPT) --- $1.2$--$1.8$pp below the claim-level default on paired differences, significant for Gemini and Claude ($+1.8$, $+1.2$) and not significant for GPT ($+1.2$, $p{=}0.075$) --- while beating every OCR condition ((C-atom)$-$(A1) $= +6.7$/$+6.3$/$+7.5$, all SIG) and the strongest protocol baseline ($+2.7$/$+3.8$/$+2.7$ vs.\ CiteEval-Auto, all SIG). Against the matched sequential chain: $+4.3$ $[1.9,6.8]$ on Gemini and $+2.8$ $[0.8,4.9]$ on Claude (both significant), $+1.1$ $[-0.8,2.9]$ (ns) on GPT --- the variant clears the chain in two families and is level in the third, where staged inference is strongest. In correction it holds the same $89.9\%$ precision as (C) at $86.5\%$ retention ($-5.3$pp; atom-level verdicts remove more correct claims), and on the human-audited naturals it detects $98.6$--$99.4\%$ of validated errors at $11.1$--$19.5\%$ false alarms --- between (C) and flat OCR on both axes. The variant remains the right tool for fine-grained auditing (atom-level verdicts, partial detection); it is not the accuracy-optimal default.

\paragraph{Why claim-level is the default (decomposition mechanism).} Decomposition is the one pipeline stage whose removal \emph{helps} ($+1.2$ to $+1.8$pp; significant for Gemini and Claude, $p{=}0.075$ for GPT), and the mechanism localizes cleanly. Splitting DocCite-Syn by the shared decomposition's atom count: on single-atom claims ($n{=}724$) the paired (C-atom)$-$(C) difference is $-0.3$/$+0.3$/$+0.6$pp (all ns) --- the two configurations are the same procedure there. On multi-atom claims ($n{=}203$) it is $-7.5$/$-6.5$/$-7.6$pp (all SIG): under strict conjunction, each atom's grounding noise compounds, and the over-flag rate on \emph{true} claims rises from $\sim$$10\%$ at one atom to $\sim$$31\%$ at two and $\sim$$79\%$ at three or more --- a family-invariant curve. Claim-level verification recovers multi-atom supported recall from $52$--$62\%$ to $89$--$92\%$ while giving back $1$--$2$pp of error sensitivity ($98.1$--$98.7\%$ to $96.8$--$97.4\%$). Decomposition remains available for fine-grained auditing (atom-level verdicts, partial detection), but it is not the accuracy-optimal default.
\paragraph{External text benchmark (sanity check only).} Running the framework's text-pair path (Gemini backbone, condition (C-atom)) on the balanced 500-item in-domain test sample of AttributionBench~\cite{attributionbench2024how} yields $70.2\%$ accuracy under the strict mapping (attributable $\Leftrightarrow$ \textsc{Verified}), with a sharply precision-skewed profile: $83.4\%$ precision but $50.4\%$ recall on the attributable class. The misses are a construct difference, not a transfer failure: false negatives concentrate in the long-form subsets (ExpertQA 67 of 124) and average $3.2$ atoms per claim, and half of them have some-but-not-all atoms verified --- the strict all-atoms conjunction reads AttributionBench's lenient holistic ``attributable'' label as partial support. Counting \textsc{Partially Verified} as attributable raises accuracy to $77.0\%$. We claim no generalization from this single configuration; it confirms only that the pipeline behaves as designed --- a strict, precision-first verifier --- on an external text benchmark it was never tuned for.

\paragraph{Model-tier $2{\times}2$ (H4).} Holding the framework fixed and swapping the backbone tier (DocCite-Syn, $n{=}928$ per cell): Gemini frontier (C-atom) $91.9\%$ / efficiency (C-atom) $91.5\%$; GPT frontier (C-atom) $91.9\%$ / efficiency (C-atom) $92.7\%$; Claude frontier (C-atom) $92.0\%$ / efficiency (C-atom) $86.0\%$. In two of three families the efficiency tier \emph{with} the framework matches or exceeds the frontier tier with the framework at roughly one-third the list price; Claude's efficiency tier does not ($-6.0$pp), so the substitution is backbone-dependent, not free. The image-over-OCR gap follows the same split: the paired (C-atom)$-$(A1) difference is $+6.7$pp (Gemini frontier, CI $[+4.5,+9.1]$), $+7.1$pp (Gemini efficiency, CI $[+4.8,+9.4]$), $+7.5$pp (GPT frontier, CI $[+5.3,+9.9]$), and $+6.9$pp (GPT efficiency, CI $[+4.3,+9.7]$) --- all significant, all $\approx{+}7$pp --- but collapses to $+0.3$pp (ns, CI $[-2.4,+3.1]$) on Claude-efficiency under the variant, whose supported-recall drops to $64.9\%$. The claim-level cell isolates the cause: under (C) the same backbone reaches $93.0\%$ binary ($86.8\%$ supported recall), $+7.1$pp $[5.2,9.0]$ over its (C-atom) cell and $+7.4$pp $[4.9,10.1]$ over flat OCR (both significant) --- the collapse is strict atom conjunction compounding the weaker backbone's grounding noise, not an inability to read the page image. Under the flagship claim-level configuration, efficiency-tier (C) reaches $93.4\%$ (Gemini), $93.0\%$ (Claude), and $93.5\%$ (GPT), matching or exceeding the frontier (C) cells at roughly one-third the list price; the tier swap remains a deployment decision to validate per family and configuration.

\section{Transfer Suite: Protocols, Anchors, and Boundary}
\label{app:transfer}

\paragraph{Protocol details.} All transfer runs use frozen prompts and greedy decoding; nonzero temperature appears only in bounded parse-retry rounds. Every reported cell finishes with zero unparsed outputs. The direct arm uses our frozen single-call JSON judge prompt in every cell, not each benchmark's native prompt. TofuEval source dialogues are reconstructed with the benchmark's released instructions from the MeetingBank transcripts and a MediaSum mirror. Llama-3-8B reads long sources in 24k-character chunks because of its 8k context. The VeriGray response-level protocol counts a summary as unfaithful when it contains at least one fabricated or contradicting sentence and drops 83 summaries whose sentences carry only gray labels, leaving 329; it is a derived protocol with no published reference row.

\paragraph{Fairness anchors.} On RAGTruth the Qwen gain is $+21.7$ F1 with 95\% CI $[18.4, 25.0]$. On TofuEval Qwen exceeds GPT-4 on the MediaSum main-topic slice (66.5 against 63.7) while GPT-4 stays well ahead on marginal topics. On VeriGray our direct arm reproduces the published 7--8B detector floor under the official sentence-level protocol (33.5 F1 against 35.9 for Lynx-8B and 35.1 for MiniCheck-7B), so the harness does not inflate or deflate baselines. One input deviation under that official protocol: both of our arms receive the document and the sentence only, without the surrounding summary.

\paragraph{Boundary conditions.} Three negative results delimit where the framework applies. (i) Pre-atomized inputs. On an LLM-AggreFact mirror pool (12{,}949 items) the direct arm reproduces the leaderboard's published Qwen2.5-7B number to 0.3 points, and the framework scores $-2.6$ $[-3.7, -1.5]$: claim-granularity input gives decomposition no room to work. (ii) Single-fact extractive answers. On 300-item stratified pilots of TRIVIA+ and HaluBench the framework drops F1 from 65.2 to 43.0 and accuracy from 75.3 to 57.6. Hallucinations there are wrong selections whose content is verbatim-present in the passage (median framework-missed response length 16 characters), so per-claim support verification passes them. (iii) Evaluation unit. On VeriGray, with the same data and the same labels, the framework loses 4.7 F1 when scored per sentence and gains 14.8 when scored per response. Together these place the framework's domain at multi-claim long-form responses; the unit of evaluation, not the dataset, sets the boundary.

\section{Algorithm, Funnel, and Qualitative Cases}

\begin{figure}[t]
\small
\begin{tabular}{@{}r@{\ \ }l@{}}
    & \textbf{Input:} answer $A$; document $D$ (pages: OCR text $+$ image) \\
 1: & $C \gets \textsc{Parse}(A)$ \hfill $\rhd$ (claim, cited page) pairs \\
 2: & \textbf{for each} $(x, p) \in C$ \textbf{in parallel:} \\
 3: & \quad $E \gets \textsc{Evidence}(D, p)$ \hfill $\rhd$ cited page $+$ $\pm2$ window \\
 4: & \quad $v_{\text{txt}} \gets \textsc{Ground}(x, \mathrm{OCR}(p))$ \hfill $\rhd$ text channel \\
    & \quad $v_{\text{img}} \gets \textsc{Ground}(x, \mathrm{Img}(p))$ \hfill $\rhd$ image channel \\
 5: & \quad $v \gets \textsc{Merge}(v_{\text{txt}}, v_{\text{img}})$ \hfill $\rhd$ hard $\to$ \textsc{inc.}; soft $\to$ image wins \\
 6: & \quad \textbf{if} $v \neq \textsc{verified}$: scan window of $E$ for support page $q$ \\
 7: & \quad\quad \textbf{if} $q$ found: $v \gets \textsc{verified-nearby}(q)$ \\
 8: & flag-only consistency check across $C$ \hfill $\rhd$ cannot alter verdicts \\
 9: & \textbf{for each} $(x, p, v) \in C$: action $\gets \textsc{Policy}(v)$ \\
    & \quad $\rhd$ \textsc{verified}$\to$\textsc{keep};\ \textsc{verified-nearby}$\to$\textsc{fix-page}$(q)$; \\
    & \quad \phantom{$\rhd$} else $\to$ \textsc{remove} \\
    & \textbf{Output:} per-citation verdicts $+$ repaired citation layer \\
\end{tabular}
\caption{Algorithm~1: AtomCite (conservative mode, claim-level default).}
\label{alg:atomcite}
\end{figure}

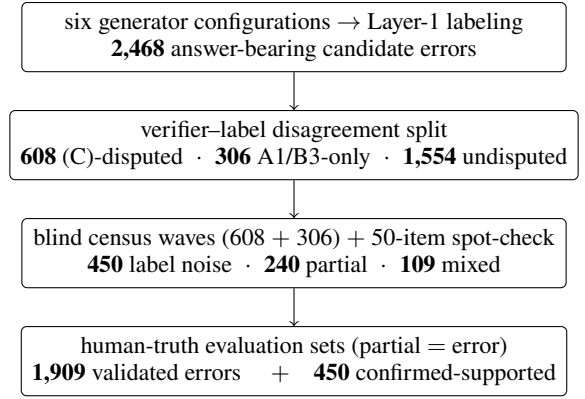
\begin{figure}[t]
\centering
\begin{tikzpicture}[
  box/.style={draw, rounded corners=2pt, align=center, font=\small, inner sep=4pt},
  lbl/.style={font=\scriptsize, midway, right=2pt},
  node distance=5mm]
\node[box, minimum width=7.2cm] (a) {six generator configurations $\to$ Layer-1 labeling\\ \textbf{2{,}468} answer-bearing candidate errors};
\node[box, minimum width=7.2cm, below=of a] (b) {verifier--label disagreement split\\ \textbf{608} (C)-disputed \;$\cdot$\; \textbf{306} A1/B3-only \;$\cdot$\; \textbf{1{,}554} undisputed};
\node[box, minimum width=7.2cm, below=of b] (c) {blind census waves (608 $+$ 306) $+$ 50-item spot-check\\ \textbf{450} label noise \;$\cdot$\; \textbf{240} partial \;$\cdot$\; \textbf{109} mixed};
\node[box, minimum width=7.2cm, below=of c] (d) {human-truth evaluation sets (partial $=$ error)\\ \textbf{1{,}909} validated errors \quad$+$\quad \textbf{450} confirmed-supported};
\draw[->] (a) -- (b); \draw[->] (b) -- (c); \draw[->] (c) -- (d);
\end{tikzpicture}
\caption{DocCite-Nat audit funnel. The spot-check found $0\%$ noise
($[0,7.3]$) in the undisputed region, whose Layer-1 labels stand.}
\label{fig:funnel}
\end{figure}

\paragraph{DocCite-Nat construction funnel.} Figure~\ref{fig:funnel} summarizes the DocCite-Nat audit funnel.

\paragraph{Qualitative failure cases.} Three recurring patterns, one real instance each (claims abridged):
\begin{itemize}
\item \emph{OCR false alarm, image rescue} (the dominant pattern behind
flat OCR's $44$--$49\%$ false-alarm rate): the claim ``Safflower
contains 78 (poly) fat in its FA composition'' is supported by a
nutrient-composition table cell on the cited page; the table's
column structure survives in the image but not in the OCR stream, so
(A1) flags a genuinely supported claim while (C) verifies it.
\item \emph{Claim-level false alarm (GPT)}: for ``The document as a
whole pertains to Youngstown State University,'' the cited first page
supports the whole-document claim only via letterhead and context;
GPT's (C) judged it unsupported --- document-scope claims anchored to a
single page are the residual false-alarm class ($12.3\%$ vs.\
$5.4$--$5.5\%$ elsewhere).
\item \emph{Chain miss (B3)}: ``The foundation named at the bottom of
the page is `NUTRITION FOUNDATION'\,'' cites a page whose footer names a
\emph{different} entity; the single-transcript chain, carrying its
earlier reasoning about neighboring pages, accepts the familiar footer
text --- state carry-over is exactly what typed per-claim sub-tasks avoid.
\end{itemize}

\section{Reproducibility and Artifacts}
All experiments use pinned model versions: \texttt{gemini-\allowbreak 3.1-\allowbreak pro-\allowbreak preview} / \texttt{gemini-\allowbreak 3-\allowbreak flash-\allowbreak preview}; \texttt{claude-\allowbreak opus-\allowbreak 4-\allowbreak 8} / \texttt{claude-\allowbreak sonnet-\allowbreak 5}; \texttt{gpt-\allowbreak 5.6-\allowbreak sol} / \texttt{gpt-\allowbreak 5.6-\allowbreak luna} (frontier / efficiency). The decomposition stage of the C-atom variant uses one pinned decomposer, shared and cached across all systems. All randomized procedures (sampling, bootstrap, annotation item order) use fixed seeds derived from a single registered constant; paired comparisons use doc-cluster bootstrap with $B{=}10^4$. All prompts are frozen verbatim in the released repository. Code, prompts, the benchmark (instances, labels, audit adjudications, annotation records), and per-run ledgers are released with the paper under licenses permitted by the source datasets (DUDE CC-BY; identifier-and-script release for MP-DocVQA). Preview-designated endpoints (the two Gemini identifiers) may be deprecated by the provider; the authors' full call ledger records every request against these pinned identifiers, so results remain traceable to the exact serving configuration.

\end{document}